%% file: main.tex
\documentclass[10pt,twocolumn,letterpaper]{article}

\usepackage[pagenumbers]{cvpr} 
\usepackage{tikz}
\usetikzlibrary{arrows.meta, positioning, calc}
\usepackage[accsupp]{axessibility}  
\usepackage{amsmath}
\usepackage{booktabs}
\usepackage{graphicx}
\definecolor{SpaceDark}{RGB}{15, 23, 42}
\definecolor{SpaceAccent}{RGB}{56, 189, 248}
\definecolor{V4Green}{RGB}{163, 230, 53}
\definecolor{V3Orange}{RGB}{249, 115, 22}
\usepackage{tikz}
\usetikzlibrary{decorations.pathreplacing}

\input{preamble}
\definecolor{cvprblue}{rgb}{0.21,0.49,0.74}
\usepackage[pagebackref,breaklinks,colorlinks,allcolors=cvprblue]{hyperref}

\def\paperID{12} 
\def\confName{CVPR}
\def\confYear{2026}

\title{Segmentation-based Detection for Efficient Multi-Task Spacecraft Perception}

\author{
Sivaperuman Muniyasamy$^{\dagger}$ \quad
Surendar Devasundaram$^{\dagger}$ \\
University of Arizona, Tucson, AZ, USA \\
\texttt{sivaastro@arizona.edu, surdev@arizona.edu} \\
{\small $\dagger$ Equal contribution}
}

\begin{document}
\maketitle
\input{sec/0_abstract}    
\input{sec/1_intro}

\input{sec/2_related_work}

\input{sec/3_approach}

\input{sec/4_experiment}

\input{sec/5_results}

\input{sec/6_conclusion}
{
    \small
    \bibliographystyle{ieeenat_fullname}
    \bibliography{main}
}

\end{document}

%% file: sec/0_abstract.tex
\begin{abstract}
Vision-based perception is fundamental to Space Situational Awareness (SSA) and autonomous on-orbit operations such as rendezvous, docking, servicing, and navigation. However, progress in this area is limited by the scarcity of annotated space imagery and by the challenging characteristics of the visual domain, including severe illumination changes, low signal-to-noise ratio, and high contrast. We address Stream 1 of the SPARK 2026 Challenge, which requires a single model for spacecraft classification, detection, and fine-grained component segmentation across multiple target types. To this end, we propose a compact hybrid architecture that integrates a MobileNet-V3 encoder with a U-Net-style decoder head, combining computational efficiency with accurate dense prediction. The resulting model is well suited to the challenge objective of developing high-performing perception systems for resource-constrained space platforms. Our method achieved an overall leaderboard score of \textbf{0.9482}, where the challenge metric is defined as \textbf{0.20Scls + 0.40Sdet + 0.40Sseg}, with task-specific scores of \textbf{1.0000} in classification, \textbf{0.9788} in detection, and \textbf{0.8917} in segmentation. The proposed approach ranked \textbf{second overall} in the SPARK 2026 Challenge, demonstrating that lightweight encoder-decoder architectures can deliver strong multi-task performance for practical onboard space vision systems. Code is available at \url{https://github.com/sivaastro/segdet-spark}.

\end{abstract}

%% file: sec/1_intro.tex
\section{Introduction}

Space Situational Awareness (SSA) is a foundational capability for safe and autonomous on-orbit operations, including inspection, rendezvous, docking, servicing, and maintenance \cite{pauly2023survey,wang2022research,li2022survey}. As orbital activity continues to increase, spacecraft must increasingly rely on onboard perception systems that can operate accurately and efficiently under strict compute, memory, and power constraints \cite{del2024mitigating}.

Vision-based spacecraft perception remains particularly challenging because of the unique characteristics of the space domain, including extreme illumination changes, high-contrast backgrounds, specular reflections, partial visibility, and large scale variation \cite{park2024robust,liu2024revisiting,zuo2024crospace6d}. These difficulties are further compounded by the scarcity of annotated real-world space imagery, which limits the direct transfer of perception models developed for terrestrial applications \cite{zhao2025automated}. Consequently, practical perception systems for space applications must strike a careful balance among robustness, accuracy, and computational efficiency.

In this work, we address Stream~1 of the SPARK~2026 Challenge, which requires a single model to perform spacecraft classification, detection, and fine-grained component segmentation under strict efficiency constraints \cite{spark2026_challenge}. Conventional multi-task perception pipelines often rely either on separate task-specific models or on detection-centric architectures with explicit bounding-box regression heads, such as Mask R-CNN- or YOLO-style designs \cite{ dung2021spacecraft,he2017mask, ren2016faster, redmon2016you}. While effective, such approaches increase architectural complexity and inference overhead, which can be undesirable in resource-constrained deployment settings. In the SPARK setting, however, each image contains a single spacecraft, making it possible to derive object localization directly from the predicted component masks rather than learning a separate bounding-box regression branch.

Motivated by this observation, we propose a lightweight multi-task architecture that combines a MobileNet-based encoder with a U-Net-style decoder \cite{howard2019searching,ronneberger2015unet}. The model predicts spacecraft component masks and class labels in a single forward pass, while bounding boxes are obtained through deterministic post-processing of the union of predicted masks. By treating segmentation as the primary dense prediction task and deriving detection from the resulting masks, the proposed design eliminates the need for a dedicated detection head and thereby reduces model complexity without sacrificing multi-task capability. Our approach achieved an overall leaderboard score of \textbf{0.9482} on the SPARK~2026 benchmark, where the official metric is defined as
\[
S_{\mathrm{acc}} = 0.20\,S_{\mathrm{cls}} + 0.40\,S_{\mathrm{det}} + 0.40\,S_{\mathrm{seg}},
\]
with task-specific scores of \textbf{1.0000} for classification, \textbf{0.9788} for detection, and \textbf{0.8917} for segmentation. This performance ranked second overall on the leaderboard, demonstrating that lightweight encoder--decoder architectures can provide a strong trade-off between accuracy and efficiency for practical multi-task spacecraft perception.

The main contributions of this work are summarized as follows:
\begin{itemize}
    \item We develop a lightweight MobileNet-based U-Net architecture for joint spacecraft classification, detection, and component segmentation under strict parameter and efficiency constraints.
    \item We show that reliable spacecraft detection can be achieved by deriving bounding boxes directly from predicted component masks, eliminating the need for a dedicated regression head in the single-object setting.    
    \item We demonstrate the effectiveness of the proposed design on the SPARK~2026 benchmark, where it achieves a leaderboard score of 0.9482 and ranks second overall.
\end{itemize}

\begin{figure*}[t]
    \centering
    \includegraphics[width=\textwidth]{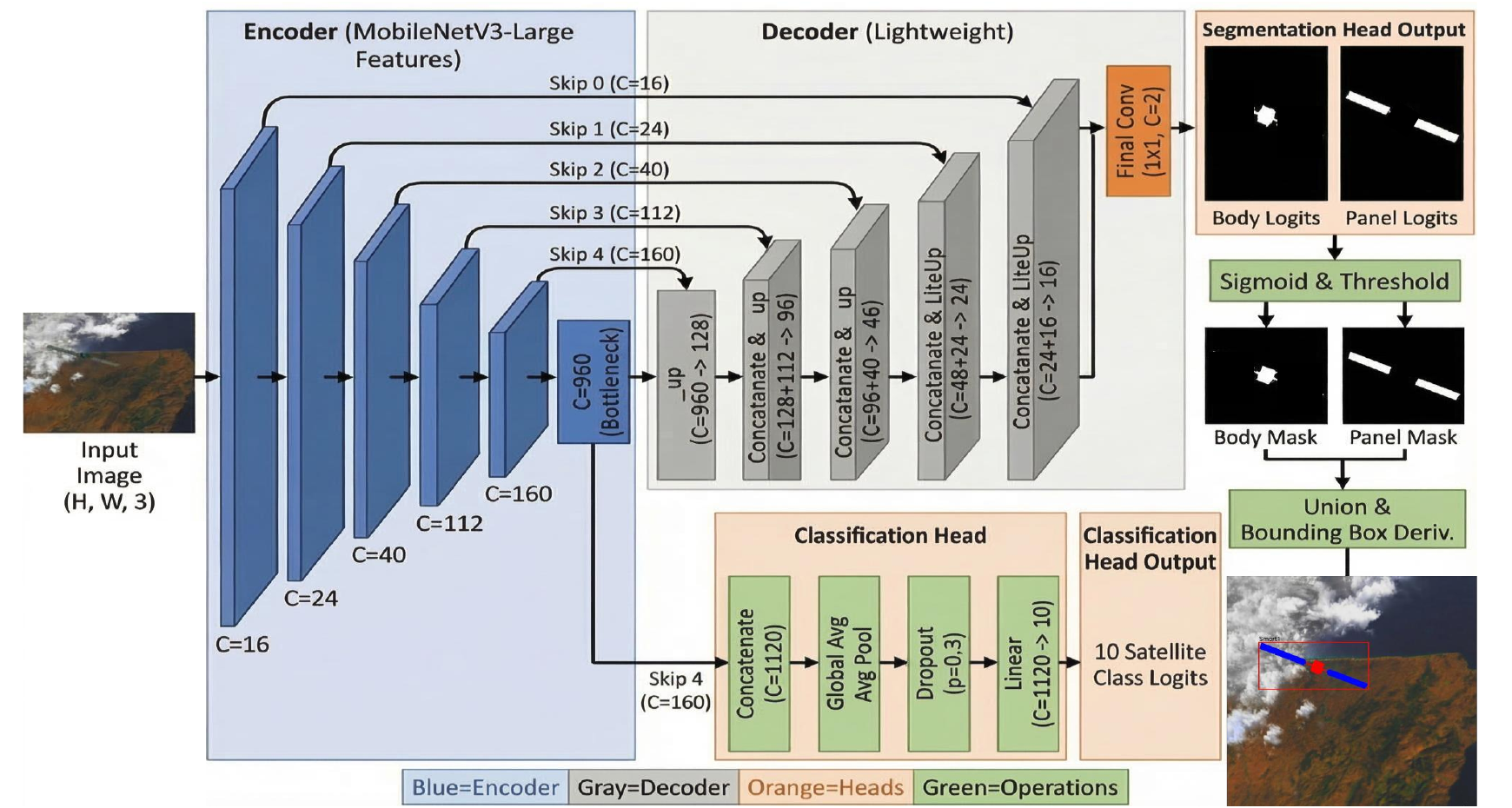}
    \caption{
    Overview of the proposed lightweight multi-task spacecraft perception architecture. A MobileNetV3-Large encoder extracts hierarchical features at multiple spatial scales, which are fused by a lightweight U-Net-style decoder through skip connections to predict two component masks corresponding to the spacecraft body and solar panels. In parallel, a compact classification head operates on deep encoder features to predict the spacecraft class. The final detection output is obtained analytically by thresholding the predicted masks, taking their union, and deriving a bounding box from the resulting foreground region.
    }
    \label{fig:architecture}
\end{figure*}

%% file: sec/2_related_work.tex
\section{Related Work}
\label{sec:related_work}

\subsection{Spacecraft perception datasets}

Progress in vision-based spacecraft perception has been strongly influenced by the availability of synthetic and hardware-in-the-loop datasets, as collecting large-scale annotated imagery in orbit remains difficult. Early benchmark datasets such as the Spacecraft PosE Estimation Dataset (SPEED) and Next Generation Spacecraft Pose Estimation Dataset (SPEED+) were introduced primarily for monocular spacecraft pose estimation and domain-gap evaluation, providing synthetic training imagery together with experimentally acquired validation data \cite{sharma2019spn,park2022speed+}. The Unreal Rendered Spacecraft On-Orbit (URSO) further demonstrated the use of Unreal Engine-based rendering for photorealistic on-orbit image generation and has been used for spacecraft vision and pose-related benchmarking \cite{proenca2019urso,zhao2023intelligent}. More recently, the Spacecraft Recognition Leveraging Knowledge (SPARK) dataset expanded the scope beyond pose estimation by providing synthetic spacecraft imagery for recognition-oriented tasks under realistic space sensing conditions \cite{musallam2021spark}. Large-scale multitask datasets have also emerged, such as the Automatically Annotated Multitask Spacecraft Dataset (AAMSD), which contains millions of automatically annotated synthetic images spanning hundreds of spacecraft and supports multiple vision tasks, including detection, semantic segmentation, and pose estimation \cite{zhao2025automated}. In addition to frame-based RGB imagery, the Event-based Observation of Spacecraft (EVOS) dataset explores event-based sensing for autonomous on-orbit inspection and related space applications \cite{evos_dataset}. Within this landscape, the SPARK~2026 Challenge focuses on efficient multi-task spacecraft perception, requiring a single model to jointly perform spacecraft classification, detection, and fine-grained component segmentation under resource-constrained deployment assumptions \cite{spark2026_challenge}.

\subsection{Models for spacecraft perception}

Existing spacecraft perception models can be broadly grouped into pose-estimation architectures, detection and instance-segmentation models, and semantic-segmentation encoder--decoder networks. In the pose-estimation setting, Spacecraft Pose Network (SPN) introduced a learning-based pipeline for monocular pose estimation on the SPEED benchmark, while SPNv2 extended this line of work through a multi-task and multi-scale formulation designed to improve robustness across domain gap \cite{sharma2019spn,park2024robust}. Although highly effective for relative pose estimation, these approaches are not designed specifically for joint part-level segmentation, object detection, and target classification.

For object detection and instance-level segmentation, detector-first architectures such as Faster R-CNN and Mask R-CNN remain standard reference points because they jointly predict class labels, bounding boxes, and segmentation masks \cite{he2017mask,ren2016faster}. However, these models typically rely on dedicated detection and regression heads, which increases architectural complexity and inference cost. Such trade-offs are important in spacecraft applications, where onboard perception systems often operate under tight memory, power, and latency constraints.

For part-level semantic understanding, encoder--decoder segmentation models are often better suited because they preserve spatial detail while producing dense predictions. U-Net introduced the now-standard skip-connected encoder--decoder design for precise localization \cite{ronneberger2015unet}. DeepLabv3+ improved dense prediction by combining multi-scale contextual aggregation with a decoder that sharpens object boundaries \cite{liu2024image}. More recently, SegFormer showed that hierarchical Transformer encoders paired with lightweight MLP decoders can achieve strong segmentation performance with competitive efficiency \cite{xie2021segformer}. In the context of spacecraft parts body: solar panel, and antenna segmentation and detection, several models are explored, such as Yolo, OCRNet, ResnetSt101,  ResnetSt200, and ASPOCNET \cite{dung2021spacecraft}. These model families are particularly relevant for spacecraft component segmentation, where thin structures such as solar panels can be lost at coarse feature resolutions.

\subsection{Efficiency-oriented multi-task perception}

The literature suggests a tension between multi-task expressiveness and deployment efficiency. Detector-centric pipelines provide explicit localization but often introduce additional heads and post-processing stages, whereas segmentation-centric encoder--decoder models provide stronger spatial reasoning with simpler dense prediction pipelines. In challenge settings such as SPARK~2026, where a single spacecraft is present in each image and fine-grained part segmentation is required, lightweight encoder--decoder architectures offer an attractive design point for jointly addressing segmentation and classification while enabling localization to be recovered from predicted masks \cite{spark2026_challenge,howard2019searching,ronneberger2015unet}. Motivated by this trade-off, our work adopts a compact MobileNet--U-Net design that emphasizes segmentation quality and derives detection through lightweight post-processing rather than through a dedicated bounding-box regression branch as shown in Figure~\ref{fig:architecture}.

%% file: sec/3_approach.tex
\section{Approach}
\label{sec:approach}

\subsection{Problem Setup}
We address the SPARK multi-task spacecraft perception problem in the single-object setting. Given an input RGB image
\begin{equation}
    I \in \mathbb{R}^{3 \times H \times W},
\end{equation}
the model jointly predicts: \emph{(i)} fine-grained segmentation of spacecraft components, \emph{(ii)} spacecraft class identity, and \emph{(iii)} object detection in the form of a bounding box. The segmentation task focuses on two foreground components, namely the spacecraft body and solar panels. Accordingly, the network outputs a two-channel segmentation logit tensor
\begin{equation}
    S \in \mathbb{R}^{2 \times H \times W},
\end{equation}
where the first channel corresponds to the body and the second channel corresponds to the solar panels. In parallel, the model predicts classification logits $c \in \mathbb{R}^{K}$,
where $K$ denotes the number of spacecraft classes.

Rather than introducing a dedicated detection head, we derive the bounding box analytically from the predicted segmentation masks. This formulation is well aligned with the SPARK setting, in which each image contains a single spacecraft. As a result, localization can be recovered directly from the segmented foreground, avoiding redundant regression of spatial information that is already encoded in the dense prediction output.

\subsection{Architecture Overview}
Our model, as shown in Figure~\ref{fig:architecture}, follows a segmentation-centric encoder--decoder design with an auxiliary classification branch. The encoder extracts hierarchical features from the input image, while the decoder progressively reconstructs high-resolution spatial detail through skip connections. The segmentation branch operates on the decoder output to predict body and panel masks, whereas the classification branch operates on compact encoder features near the bottleneck. The final detection output is then obtained by deterministic post-processing of the predicted masks.

This design is motivated by two requirements of the target setting. First, spacecraft component segmentation requires precise recovery of thin structures such as solar panels and appendages, which benefits from multi-scale skip connections and dense prediction. Second, onboard deployment imposes strict efficiency constraints, which favor lightweight backbones and inexpensive decoding operations. Our architecture therefore combines a mobile-scale encoder with a compact U-Net-style decoder and avoids task-specific heads that would substantially increase parameter count and inference cost.

\subsection{MobileNetV3 Encoder}
To achieve favorable accuracy--efficiency trade-offs, we adopt MobileNetV3-Large as the encoder backbone. MobileNetV3 is built from inverted residual bottleneck blocks with depthwise separable convolutions, lightweight channel attention through squeeze-and-excitation modules, and efficient nonlinearities such as hard-swish. These design choices make it well suited for high-resolution dense prediction under constrained compute budgets.

The encoder progressively reduces spatial resolution while increasing semantic abstraction. For an input image of size $H \times W$, skip features are extracted at four spatial scales,
\begin{equation}
    \frac{H}{2},\quad \frac{H}{4},\quad \frac{H}{8},\quad \frac{H}{16},
\end{equation}
while the deepest bottleneck representation is computed at
\begin{equation}
    \frac{H}{32}.
\end{equation}
For an input resolution of $1024 \times 1024$, the bottleneck feature map has spatial size $32 \times 32$ and 960 channels in the MobileNetV3-Large configuration. These hierarchical features provide complementary information: shallow layers preserve fine boundary and texture cues, whereas deeper layers encode higher-level semantic context. The decoder exploits both through skip connections to improve the localization of thin and elongated spacecraft components.

\subsection{Lite Decoder}
Starting from the encoder bottleneck, the decoder progressively upsamples the feature maps back to the input resolution. A straightforward decoder with uniformly wide channels and transposed convolutions at all scales becomes expensive at high spatial resolutions, especially for $1024 \times 1024$ inputs. To reduce this cost, we employ a lightweight decoder with tapered channel widths and inexpensive upsampling operations in the final stages.

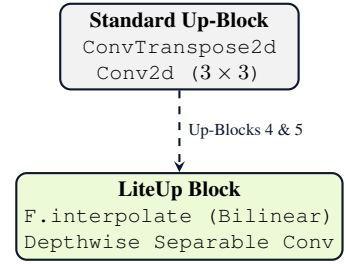
\begin{figure}[t]
\centering
\begin{tikzpicture}[
    >=stealth,
    scale=0.9,
    transform shape,
    box/.style={
        draw=SpaceDark,
        thick,
        rounded corners,
        fill=white,
        align=center,
        minimum width=3.5cm,
        minimum height=1cm,
        font=\small
    },
    arr/.style={->, thick, SpaceDark}
]
    \node[box, fill=gray!10] (std) at (0,0) {
        \textbf{Standard Up-Block}\\
        \texttt{ConvTranspose2d}\\
        \texttt{Conv2d ($3\times3$)}
    };

    \node[box, fill=V4Green!20] (lite) at (0,-2.5) {
        \textbf{LiteUp Block}\\
        \texttt{F.interpolate (Bilinear)}\\
        \texttt{Depthwise Separable Conv}
    };

    \draw[arr, dashed] (std) -- node[right, font=\scriptsize] {Up-Blocks 4 \& 5} (lite);
\end{tikzpicture}
\caption{Replacement of the final high-resolution upsampling blocks with the proposed LiteUp design.}
\label{fig:liteup_block}
\end{figure}

Specifically, the decoder channel widths follow the schedule
\begin{equation}
    128 \rightarrow 96 \rightarrow 48 \rightarrow 24 \rightarrow 16.
\end{equation}
The early upsampling stages operate on relatively small spatial maps and use standard learned upsampling blocks. In contrast, the final high-resolution stages replace transposed convolutions with a lighter design consisting of bilinear interpolation followed by depthwise separable convolution, as shown in Figure~\ref{fig:liteup_block}. This choice reduces the computational burden where the spatial resolution is largest while still allowing local refinement after upsampling.

The resulting decoder preserves the strengths of a U-Net-style architecture---namely, multi-scale fusion and accurate boundary recovery---while keeping the parameter count and FLOP budget suitable for resource-constrained deployment. In particular, the combination of skip connections and lightweight refinement blocks helps recover fine structures that are easily degraded by aggressive downsampling in the encoder. It achieves strict competition efficiency constraints without compromising the fine-grained localization needed for solar panel segmentation, reducing the overall footprint from approximately $58.13$ GFLOPs in the standard decoder formulation to just $22.575$ GFLOPs.

\subsection{Multi-Task Heads and Mask-Derived Detection}
\paragraph{Segmentation head.}
The decoder produces a two-channel segmentation logit map $S$, corresponding to the body and solar panel regions. A sigmoid activation is applied independently to each channel, yielding per-pixel foreground probabilities for each component class. This formulation allows the model to focus on part-level delineation rather than only coarse object extent, which is important for spacecraft imagery containing thin appendages and sharp structural boundaries.

\paragraph{Classification head.}
Spacecraft identity is predicted from the compact encoder representation at the bottleneck. We apply global pooling to the deepest feature map concatenated with skip 4 and feed the resulting feature vector to a lightweight linear classification layer, producing class logits $c \in \mathbb{R}^{K}$. This branch introduces minimal overhead while leveraging semantically rich encoder features for global recognition.

\paragraph{Mask-derived detection.}
Instead of learning a separate bounding-box regression head, we derive the detection output from the predicted segmentation masks. Let $S_b$ and $S_p$ denote the body and panel logits, respectively. After sigmoid activation and thresholding, the binary component masks are
\begin{equation}
M_b =
\begin{cases}
1, & \sigma(S_b) > \tau_b,\\
0, & \text{otherwise},
\end{cases}
\qquad
M_p =
\begin{cases}
1, & \sigma(S_p) > \tau_p,\\
0, & \text{otherwise}.
\end{cases}
\label{eq:mask_threshold}
\end{equation}
where $\tau_b$ and $\tau_p$ are foreground thresholds and $\sigma(\cdot)$ denotes the sigmoid function. We then define the overall spacecraft foreground mask as the union
\begin{equation}
    M = M_b \cup M_p.
\end{equation}
The final detection box is the tight axis-aligned bounding rectangle enclosing the active pixels in $M$, optionally enlarged by a small fractional padding factor $\delta$ to account for boundary uncertainty. This yields a deterministic localization rule that is simple, parameter-free, and naturally coupled to the segmentation output.

The key advantage of this formulation is that it removes the need for a dedicated detection head in the single-object setting. Consequently, the model can allocate its capacity to the two dense tasks that matter most for this benchmark---component segmentation and class prediction---while still producing accurate bounding boxes through post-processing.

\subsection{Training Objective}
We jointly optimize the segmentation and classification tasks. The segmentation objective combines complementary region- and boundary-aware losses:
\begin{equation}
    \mathcal{L}_{\mathrm{seg}}
    =
    \lambda_{\mathrm{bce}} \mathcal{L}_{\mathrm{BCE}}
    +
    \lambda_{\mathrm{dice}} \mathcal{L}_{\mathrm{Dice}}
    +
    \lambda_{\mathrm{lovasz}} \mathcal{L}_{\mathrm{Lovasz}}.
\end{equation}
In our implementation, the loss weights are set to
\begin{equation}
    \lambda_{\mathrm{bce}} = 0.3,\qquad
    \lambda_{\mathrm{dice}} = 0.3,\qquad
    \lambda_{\mathrm{lovasz}} = 0.4.
\end{equation}
Binary cross-entropy provides stable pixel-wise supervision, Dice loss improves overlap quality for imbalanced foreground regions, and Lov\'asz loss encourages better optimization of mask quality near object boundaries. We apply Dice loss independently to each foreground channel so that the comparatively thin solar-panel regions are not dominated by the larger body mask.

The classification branch is supervised with standard cross-entropy:
\begin{equation}
    \mathcal{L}_{\mathrm{cls}} = \mathcal{L}_{\mathrm{CE}}(c, y),
\end{equation}
where $y$ denotes the ground-truth spacecraft class label. The overall training objective is
\begin{equation}
    \mathcal{L}
    =
    \mathcal{L}_{\mathrm{seg}}
    +
    \lambda_{\mathrm{cls}} \mathcal{L}_{\mathrm{cls}},
\end{equation}
with $\lambda_{\mathrm{cls}} = 0.1$ in our experiments. This weighting emphasizes accurate component segmentation while still encouraging discriminative global features for spacecraft classification.

%% file: sec/4_experiment.tex
\section{Experiments}
\label{sec:experiments}

\subsection{Experimental Setup}
\label{sec:experimental_setup}

\begin{table}[t]
\centering
\setlength{\tabcolsep}{4pt}
\caption{Overall performance and efficiency on SPARK~2026 Stream-1 for our MobileNetV3-Large + U-Net-style decoder and the SegFormer-B0 baseline.}
\label{tab:overall_results}
\begin{tabular}{lcccc}
\toprule
Model & $S_{acc}$ & $S_{final}$ & GFLOPs & Params \\
\midrule
Ours         & 0.9482 & 0.9276 & 22.575 & 3.849M \\
SegFormer-B0 & 0.9351 & 0.8917 & 30.413 & 3.454M \\
\bottomrule
\end{tabular}
\end{table}
\begin{table}[t]
\centering
\caption{Task-wise performance on SPARK~2026 Stream-1 for our MobileNetV3-Large + U-Net-style decoder and the SegFormer-B0 baseline..}
\label{tab:task_breakdown}
\begin{tabular}{lccc}
\toprule
Model & $S_{cls}$ & $S_{det}$ & $S_{seg}$ \\
\midrule
Ours & 1.0000 & 0.9788 & 0.8917 \\
SegFormer-B0 & 1.0000 & 0.9439 & 0.8939 \\
\bottomrule
\end{tabular}
\end{table}
\begin{table}[t]
\centering
\scriptsize
\setlength{\tabcolsep}{3pt}
\caption{Impact of foreground threshold on validation performance.}
\label{tab:threshold_ablation}
\resizebox{\columnwidth}{!}{%
\begin{tabular}{@{}lccccccc@{}}
\toprule
$\tau$ & $S_{\mathrm{cls}}$ & $S_{\mathrm{det}}$ & Body & Panel & $S_{\mathrm{seg}}$ & $S_{\mathrm{acc}}$ & $S_{\mathrm{final}}$ \\
\midrule
0.30 & 0.9998 & \textbf{0.9601} & 0.9148 & 0.8559 & 0.9050 & 0.9460 & 0.9091 \\
0.35 & 0.9998 & 0.9597 & 0.9225 & 0.8664 & 0.9128 & 0.9490 & 0.9112 \\
0.40 & 0.9998 & 0.9589 & 0.9266 & 0.8727 & 0.9173 & 0.9505 & 0.9122 \\
0.45 & 0.9998 & 0.9578 & 0.9286 & 0.8765 & 0.9197 & \textbf{0.9510} & \textbf{0.9126} \\
0.50 & 0.9998 & 0.9565 & \textbf{0.9295} & \textbf{0.8785} & \textbf{0.9209} & 0.9509 & 0.9125 \\
0.55 & 0.9998 & 0.9510 & 0.9288 & 0.8778 & 0.9203 & 0.9485 & 0.9108 \\
0.60 & 0.9998 & 0.9480 & 0.9266 & 0.8753 & 0.9181 & 0.9464 & 0.9094 \\
\bottomrule
\end{tabular}%
}
\end{table}

We evaluate our method on the \textbf{SPARK~2026 Stream-1} benchmark, which jointly scores spacecraft \emph{classification}, \emph{detection}, and \emph{fine-grained segmentation} under efficiency-aware deployment constraints. Following the official challenge protocol, the task accuracy score is defined as
\begin{equation}
S_{\mathrm{acc}} = 0.20 S_{\mathrm{cls}} + 0.40 S_{\mathrm{det}} + 0.40 S_{\mathrm{seg}},
\end{equation}
where $S_{\mathrm{cls}}$, $S_{\mathrm{det}}$, and $S_{\mathrm{seg}}$ denote the classification, detection, and segmentation scores, respectively.

The final competition score additionally incorporates computational efficiency:
\begin{equation}
S_{\mathrm{final}} = 0.70 S_{\mathrm{acc}} + 0.15 S_{\mathrm{flops}} + 0.15 S_{\mathrm{params}},
\end{equation}
where
\begin{equation}
S_{\mathrm{flops}} = \max\left(0, 1 - \frac{\mathrm{GFLOPs}}{100}\right),
\end{equation}
and
\begin{equation}
S_{\mathrm{params}} = \max\left(0, 1 - \frac{\mathrm{MParams}}{30}\right).
\end{equation}

This protocol rewards both predictive performance and deployability, making it particularly relevant for onboard spacecraft perception under resource-constrained settings.

\subsection{Implementation Details}
\label{sec:implementation_details}

All models operate on RGB images at a spatial resolution of $1024 \times 1024$. Our primary model uses the MobileNetV3-Large encoder and lightweight U-Net-style decoder described in Sec.~\ref{sec:approach}. The network is trained jointly for segmentation and classification using the composite objective defined in Sec.~\ref{sec:approach}.

To adapt pretrained encoder features while preserving efficient learning of task-specific layers, we use differential learning rates: $10^{-4}$ for the encoder and $10^{-3}$ for the decoder and task heads. Training is performed with Distributed Data Parallel (DDP) across multiple GPUs. We further employ automatic mixed precision to accelerate high-resolution training, using BF16 on A100/H100 hardware and FP16 otherwise. Gradient accumulation is used to maintain a stable effective batch size during optimization.

At inference time, detection boxes are derived analytically from the union of the predicted body and panel masks. Unless otherwise stated, the derived box is expanded by a relative padding factor of $2.5\%$, which improves localization robustness while introducing no additional parameters or learnable computation.

%% file: sec/5_results.tex
\section{Results}
\label{sec:results}

\subsection{Main Quantitative Results}
\label{sec:main_results}

Our \textbf{MobileNetV3-Large + U-Net-style decoder} achieves a task accuracy score of \textbf{\emph{$S_{\mathrm{acc}} = 0.9482$}}, ranking \textbf{second on the SPARK~2026 Stream-1 leaderboard}. The model requires only \textbf{22.575 GFLOPs} and \textbf{3.849M parameters} (computed using fvcore library), yielding \textbf{\emph{$S_{\mathrm{flops}} = 0.8871$}} and \textbf{\emph{$S_{\mathrm{params}} = 0.8717$}}, with a final competition score of \textbf{\emph{$S_{\mathrm{final}} = 0.9276$}}. Task-wise, the model performs strongly across all outputs, with perfect classification, high detection quality, and competitive fine-grained segmentation. These results show that a lightweight segmentation-centered architecture can remain highly competitive under an efficiency-aware benchmark.

\begin{table}[t]
\centering
\scriptsize
\setlength{\tabcolsep}{3pt}
\caption{Impact of box padding on validation performance for mask-derived detection.}
\label{tab:padding_ablation}
\resizebox{\columnwidth}{!}{%
\begin{tabular}{@{}lccccccc@{}}
\toprule
$p$ & $S_{\mathrm{cls}}$ & $S_{\mathrm{det}}$ & Body & Panel & $S_{\mathrm{seg}}$ & $S_{\mathrm{acc}}$ & $S_{\mathrm{final}}$ \\
\midrule
0     & 0.9998 & 0.8425 & 0.9295 & 0.8785 & 0.9209 & 0.9053 & 0.8806 \\
1.0\% & 0.9998 & 0.8985 & 0.9295 & 0.8785 & 0.9209 & 0.9277 & 0.8963 \\
1.5\% & 0.9998 & 0.9270 & 0.9295 & 0.8785 & 0.9209 & 0.9391 & 0.9043 \\
2.0\% & 0.9998 & 0.9462 & 0.9295 & 0.8785 & 0.9209 & 0.9468 & 0.9097 \\
2.5\% & 0.9998 & 0.9565 & 0.9295 & 0.8785 & 0.9209 & 0.9509 & 0.9125 \\
3.0\% & 0.9998 & \textbf{0.9589} & 0.9295 & 0.8785 & 0.9209 & \textbf{0.9519} & \textbf{0.9132} \\
4.0\% & 0.9998 & 0.9275 & 0.9295 & 0.8785 & 0.9209 & 0.9393 & 0.9044 \\
\bottomrule
\end{tabular}%
}
\end{table}

\begin{table*}[t]
\centering
\caption{Decoder ablation on the MobileNetV3 backbone. Replacing LiteUp with a wider standard decoder degrades performance while increasing model size and computation. LiteUp is therefore used in the final model.}
\label{tab:decoder_ablation}
\begin{tabular}{lccccccccc}
\toprule
Decoder & $S_{\mathrm{cls}}$ & $S_{\mathrm{det}}$ & Body & Panel & $S_{\mathrm{seg}}$ & $S_{\mathrm{acc}}$ & Params (M) & GFLOPs & $S_{\mathrm{final}}$ \\
\midrule
LiteUp (ours)      & \textbf{0.9998} & \textbf{0.9565} & \textbf{0.9295} & \textbf{0.8785} & \textbf{0.9209} & \textbf{0.9509} & \textbf{3.849} & \textbf{22.575} & \textbf{0.9125} \\
Standard decoder   & 0.8352 & 0.7532 & 0.7465 & 0.6515 & 0.7368 & 0.7630 & 5.011 & 58.130 & 0.7219 \\
\bottomrule
\end{tabular}
\end{table*}

\subsection{Comparison with SegFormer-B0}
\label{sec:segformer_comparison}

We compare our lightweight CNN-based design with the transformer-based \textbf{SegFormer-B0}. SegFormer-B0 attains slightly better segmentation performance ($S_{\mathrm{seg}} = 0.8939$ vs.\ $0.8917$), but at substantially higher computational cost: \textbf{30.413 GFLOPs} versus \textbf{22.575 GFLOPs}, with comparable parameter counts (\textbf{3.454M} vs.\ \textbf{3.849M}). Tables~\ref{tab:overall_results} and~\ref{tab:task_breakdown} summarize the comparison.

Under the official metric, this higher compute cost reduces the efficiency terms to \textbf{\emph{$S_{\mathrm{flops}} = 0.6959$}} and \textbf{\emph{$S_{\mathrm{params}} = 0.8849$}}, leading to a final competition score of \textbf{\emph{$S_{\mathrm{final}} = 0.8917$}}. Although SegFormer-B0 produces slightly stronger masks, our MobileNetV3-Large model delivers substantially better detection ($0.9788$ vs.\ $0.9439$) and a better overall efficiency-accuracy trade-off, resulting in a higher final score. This highlights that in efficiency-aware spacecraft perception benchmarks, small segmentation gains may not translate into better overall performance when accompanied by a large FLOP penalty.

\subsection{Ablation Study}
\label{sec:mask_detection_results}
\indent\textbf{Foreground Threshold sensitivity} Table~\ref{tab:threshold_ablation} shows that the model is fairly robust to the foreground threshold over a broad range. While lower thresholds slightly improve detection, higher thresholds improve mask quality by suppressing false positives. The best overall validation score is obtained at $\tau=0.5$, which provides the most favorable balance between detection and segmentation and is used in the final configuration.

\indent\textbf{Padding sensitivity}
Our method derives the detection box from the union of the predicted body and panel masks instead of learning a separate box-regression head. This is well suited to the benchmark, since each image contains a single spacecraft and segmentation is already required. However, incomplete masks, especially around thin structures such as solar panels, can slightly underestimate the derived box. To address this, we enlarge the box with a fixed relative padding factor of $3.0\%$, selected from a validation sweep.
Table~\ref{tab:padding_ablation} shows that this simple post-processing step leaves classification and segmentation unchanged, while substantially improving detection from $0.8425$ to $0.9589$. As a result, the overall score increases from $0.8806$ to $0.9132$. This demonstrates that lightweight geometric post-processing can significantly improve localization robustness without adding parameters or FLOPs.

\indent\textbf{Lite Decoder vs Standard Decoder} Table~\ref{tab:decoder_ablation} shows that the proposed LiteUp decoder achieves the best overall trade-off between accuracy and efficiency. Compared with the standard decoder variant, LiteUp yields substantially stronger classification, detection, and segmentation performance while also requiring fewer parameters and much lower computation. These results support the use of LiteUp in the final model.

\indent\textbf{Loss weight sensitivity}
We analyzed the sensitivity of the multi-task objective using Optuna. The top-performing trials consistently favored a \emph{small classification loss weight} and a \emph{dominant} segmentation objective. The best trial achieved a validation score of $0.9509$ with $\lambda_{\mathrm{cls}}=0.1$, $\lambda_{\mathrm{bce}}=0.3$, $\lambda_{\mathrm{dice}}=0.3$, and $\lambda_{\mathrm{lovasz}}=0.4$. More broadly, the top-5 trials are shown in the Table~\ref{tab:optuna_top5}, where performance was strongest when the segmentation loss emphasized stable pixel-wise supervision, while the classification branch remained lightly weighted as an auxiliary task.
\begin{table}
\centering
\caption{Top-5 Optuna trials for loss-weight sensitivity.}
\label{tab:optuna_top5}
\resizebox{\columnwidth}{!}{%
\begin{tabular}{c c c c c c}
\toprule
Trial & Val. Score & $\lambda_{\mathrm{cls}}$ & $\lambda_{\mathrm{bce}}$ & $\lambda_{\mathrm{dice}}$ & $\lambda_{\mathrm{lovasz}}$ \\
\midrule
62 & \textbf{0.9509} & \textbf{0.1} & \textbf{0.3} & \textbf{0.3} & \textbf{0.4} \\
39 & 0.7118 & 0.0328 & 0.7182 & 0.1146 & 0.1672 \\
38 & 0.6849 & 0.0343 & 0.7184 & 0.1156 & 0.1659 \\
42 & 0.6799 & 0.0405 & 0.5412 & 0.2144 & 0.2443 \\
19 & 0.6754 & 0.0533 & 0.7006 & 0.1126 & 0.1869 \\
\bottomrule
\end{tabular}%
}
\end{table}

\subsection{Discussion}
\label{sec:discussion}

The results suggest that a \textbf{segmentation-centered multi-task architecture} is an effective design choice for efficiency-aware spacecraft perception. By combining a lightweight mobile encoder, a compact decoder, and analytic mask-derived detection, our model achieves a strong balance between segmentation fidelity, localization quality, and computational efficiency.

More broadly, these findings indicate that the best model for deployment-oriented spacecraft perception is not necessarily the one with the highest raw segmentation score, but rather the one that achieves the most favorable trade-off among dense prediction quality, localization consistency, and resource usage. This observation is particularly relevant for autonomous spacecraft systems, where perception must operate reliably within tight onboard hardware budgets.

%% file: sec/6_conclusion.tex
\section{Conclusion}
\label{sec:conclusion}

In this work, we presented a lightweight multi-task perception model for SPARK~2026 Stream-1, targeting spacecraft classification, detection, and fine-grained component segmentation within a single architecture. Our approach combines a MobileNetV3 encoder with a compact U-Net-style decoder, using segmentation as the primary dense prediction task and deriving detection analytically from the union of predicted component masks. This design removes the need for a dedicated bounding-box regression head, reduces architectural complexity, and remains well aligned with the single-spacecraft setting of the benchmark.

The proposed model achieved a task accuracy score of 0.9482 and ranked second on the SPARK~2026 Stream-1 leaderboard, while requiring only 22.575 GFLOPs and 3.849M parameters. The results show that strong multi-task performance can be achieved without resorting to heavier detection-centric or transformer-based designs. In particular, our experiments demonstrate that efficient encoder--decoder architectures, combined with lightweight geometric post-processing, provide a favorable balance among segmentation quality, localization accuracy, and deployment efficiency.

More broadly, this work highlights the value of segmentation-centered formulations for resource-aware spacecraft perception. Rather than treating detection, classification, and segmentation as entirely separate objectives with independent task heads, our results suggest that dense part-level prediction can serve as a compact and informative representation from which other outputs can be recovered. This is especially relevant for autonomous onboard systems, where perception models must operate under strict memory, compute, and power constraints.

As future work, the proposed framework could be extended beyond the single-object setting to more general multi-spacecraft scenes, where learned instance separation or more structured localization mechanisms may be required. Further investigation into robustness under real imagery, sim-to-real transfer, and temporally consistent perception for navigation and proximity operations would also strengthen the applicability of lightweight multi-task spacecraft vision systems in practical space missions.